%% file: iclr2021_conference.tex
\documentclass{article} % For LaTeX2e
\usepackage{iclr2021_conference,times}

\input{math_commands.tex}

\usepackage{hyperref}
\usepackage{url}
\usepackage{graphicx}
\usepackage{caption}
\usepackage{subcaption}
\usepackage[export]{adjustbox}
\usepackage{algorithm,algpseudocode}
\usepackage{todonotes}
\usepackage{wrapfig}
\usepackage{amsmath}
\usepackage{multirow}
\usepackage{multicol}
\usepackage{array}
\usepackage[export]{adjustbox}

\title{Adversarial Environment Generation for\\ Learning to Navigate the Web}

\author{Izzeddin Gur, Natasha Jaques, Kevin Malta, Manoj Tiwari, Honglak Lee, Aleksandra Faust \\
Google Research, Mountain View, CA, 94043 \\
\texttt{\{izzeddin,natashajaques,kmalta,mjtiwari,honglak,sandrafaust\}@google.com}
}

\newcommand{\adversary}{adversary}
\newcommand{\webnav}{navigator}

\newcommand{\WEDP}{Flexible PAIRED}
\newcommand{\WEDPb}{Flexible b-PAIRED}

\iclrfinalcopy % Uncomment for camera-ready version, but NOT for submission.
\begin{document}
\maketitle

\begin{abstract}
Learning to autonomously navigate the web is a difficult sequential decision-making task. 
The state and action spaces are large and combinatorial in nature, and websites are dynamic environments consisting of several pages.
One of the bottlenecks of training web navigation agents is providing a learnable curriculum of training environments that can cover the large variety of real-world websites.
Therefore, we propose using Adversarial Environment Generation (AEG) to generate challenging web environments in which to train reinforcement learning (RL) agents. We provide a new benchmarking environment, gMiniWoB, which enables an RL \adversary\ to use compositional primitives to learn to generate arbitrarily complex websites. To train the adversary, we propose a new technique for maximizing regret using the difference in the scores obtained by a pair of \webnav\ agents. Our results show that our approach significantly outperforms prior methods for minimax regret AEG. The regret objective trains the \adversary\ to design a curriculum of environments that are ``just-the-right-challenge'' for the \webnav\ agents; our results show that over time, the \adversary\ learns to generate increasingly complex web navigation tasks. The \webnav\ agents trained with our technique learn to complete challenging, high-dimensional web navigation tasks, such as form filling, booking a flight etc.
We show that the \webnav\ agent trained with our proposed \WEDPb\ technique significantly outperforms competitive automatic curriculum generation baselines---including a state-of-the-art RL web navigation approach---on a set of challenging unseen test environments, and achieves more than 80\% success rate on some tasks.
\end{abstract}

\section{Introduction}

The goal of this work is to train reinforcement learning (RL) agents to navigate the web; specifically, by correctly entering relevant information into unknown, real-world websites. This ability could enable a user to issue requests such as, ``Buy me a plane ticket to Los Angeles leaving on Friday'', or ``Post the following on my social media account'', and have the RL agent automatically handle the details of completing these tasks. However, the complexity and diversity of real-world websites makes this a formidable challenge.

To enable our agents to generalize to novel websites, they operate directly on the Document Object Model (DOM). The DOM is a tree of web elements, and agents must correctly select and fill out the appropriate elements. This makes the state-action space of the problem prohibitively large. Even if the agent is able to navigate the site to arrive at the correct form, and eventually select the correct element (e.g. the `departure' field for booking a flight), there are many possible values it can insert (e.g. all user input). To mitigate this issue, past work  \citep{shi2017world,liu2018reinforcement} has relied on behavior cloning from expert demonstrations. However, this approach is brittle and cannot scale effectively.
It is not possible to obtain demonstrations for navigating every possible website, especially since sites are frequently changed and updated.
If there is no demonstration data available, a model based on imitation learning is unlikely to be able to generalize to a novel website. 

Successfully navigating the wide range of real-world websites requires training an agent on a large distribution of possible tasks and environments. The question is how to create a distribution that will not only cover most real-world tasks, but can be presented in a curriculum that is learnable by the agent. One option would be to manually design a pre-defined curriculum of hand-built websites. However, this is tedious, time-consuming, error-prone, and brittle; the designer is likely to miss some real-world edge cases. Another option would be to apply domain randomization (DR) (as in e.g. \citet{jakobi1997evolutionary,sadeghi2016cad2rl,tobin2017domain}) to randomize parameters of websites, or automatically increase some parameter controlling the difficulty over time (as in \citet{gur2018learning}). However, both approaches may fail to cover important test cases, and cannot tailor the difficulty of the parameter configuration to the current ability of the agent.

\begin{figure}
\vspace{-1.cm}
    \center
    \subfloat[Early training]{
        \includegraphics[width=0.23\linewidth]{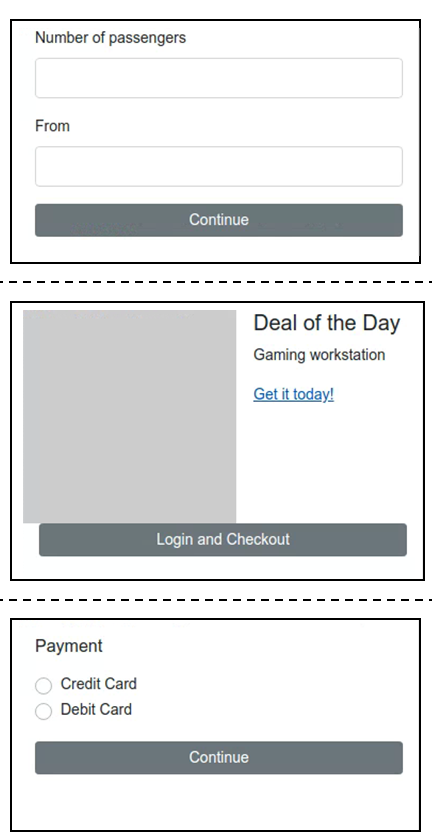}}
    \subfloat[Mid training]{
        \includegraphics[width=0.23\linewidth]{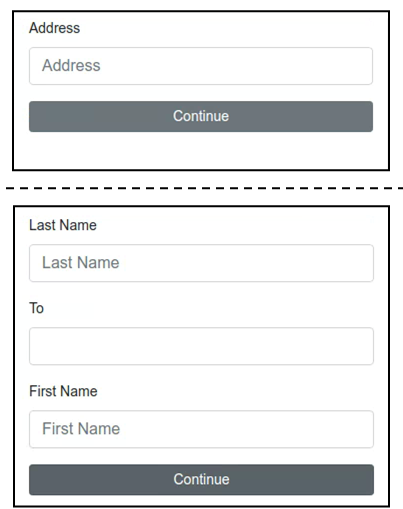}}
    \subfloat[Late training]{
        \includegraphics[width=0.23\linewidth,fbox]{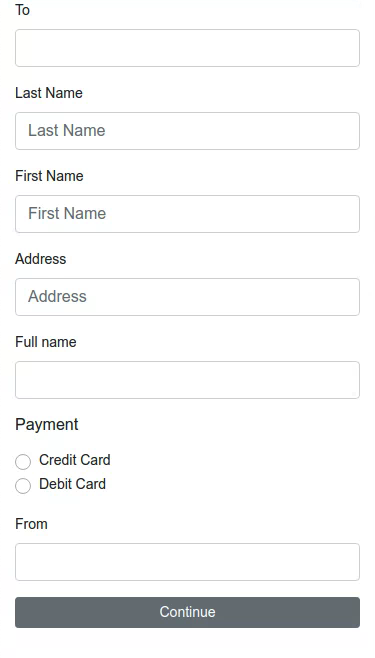}}
    \subfloat[Test]{
        \includegraphics[width=0.23\linewidth,fbox]{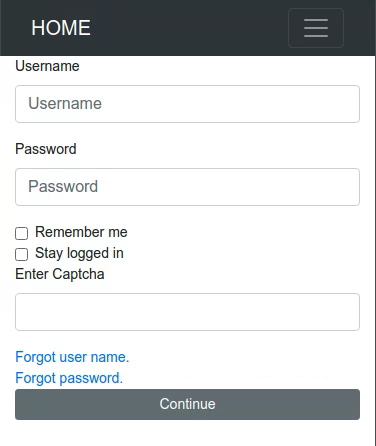}}
    \caption{\small Samples of generated web pages from selected websites taken from early, middle, and late snapshots of the training (a-c) and unseen test ``Login'' website (d). Over time, the number of pages in a website decreases but the density of elements in a page increases with more task-oriented elements.}
    \label{fig:samplewebsites}
    \vspace{-0.75cm}
\end{figure}

Therefore, in this work we leverage cutting-edge techniques for Adversarial Environment Generation (AEG) to build a curriculum of challenging web navigation tasks. Specifically, we train an adversarial RL agent to learn to create new pages in a web site in order to exploit the current weaknesses in an agent that is learning to navigate the web. To enable this AEG web-design technique, we build a new framework, gMiniWoB, that enables an adversary to construct websites out of common design primitives such as \textit{navigation bars}, \textit{product carousels}, \textit{item decks}, \textit{web forms}, and \textit{item carts}. We are releasing this environment in open-source in the hopes of enabling further progress on this problem. To the best of our knowledge, we are the first to apply AEG to web navigation. 

The goal of AEG is to automatically generate a curriculum of training environments that will cover the space of possible websites, and thereby enable generalization to real-world web navigation tasks. However, if we naively apply a minimax adversary---i.e. an adversary that seeks to minimize the performance of the learning agent---this curriculum is unlikely to emerge. This is because the adversary is motivated to create the hardest possible website, rather than tailor the difficulty of the site to the current skill level of the agent. Instead, PAIRED (Protagonist Antagonist Induced Regret Environment Design) \citep{dennis2020emergent}, a recently proposed AEG technique, trains the adversary to maximize the \textit{regret}. 
We improve upon the original PAIRED algorithm with two novel algorithmic enhancements. First, we propose a more flexible method for computing the regret which makes our algorithm less vulnerable to becoming stuck in a local minimum. Second, we introduce an explicit budgeting mechanism, such that the adversary is penalized for making more complex environments when the agents cannot solve the task, and otherwise rewarded for making complex environments.

This paper makes the following contributions: i) A new benchmarking environment, gMiniWoB, which empowers the use of Adversarial Environment Generation for web navigation, by enabling the construction of websites out of compositional design primitives; ii) The \WEDPb\ algorithm, which computes a more stable estimate of regret and directly incentivizes the adversary to tailor the complexity of the generated environment to the performance of the agent; and iii) empirical results demonstrating that \WEDPb\ generates a curriculum of increasingly challenging websites, and produces agents that can successfully generalize to navigating complex, unseen sites at test time. Our approach significantly outperforms prior work on minimax regret AEG \citep{dennis2020emergent}, as well as a state-of-the-art approach for using RL to train web navigation agents \citep{gur2018learning}.  We hope that this work will provide a meaningful way to make progress on the exceptionally challenging problem of learning to navigate the web, and will be of interest to the wider RL research community for auto-curriculum design in complex and compositional environments.

\input{src/related}
\input{src/background}
\input{src/model}
\input{src/experiments}
\input{src/conclusion}

\bibliography{iclr2021_conference}
\bibliographystyle{iclr2021_conference}

\appendix
\input{src/appendix}

\end{document}

%% file: math_commands.tex
\usepackage{amsmath,amsfonts,bm}

\def\eqref#1{equation~\ref{#1}}
\def\1{\bm{1}}

\DeclareMathAlphabet{\mathsfit}{\encodingdefault}{\sfdefault}{m}{sl}
\SetMathAlphabet{\mathsfit}{bold}{\encodingdefault}{\sfdefault}{bx}{n}

%% file: src/related.tex
\section{Related work}
Prior work on training agents to navigate the web introduced the Miniwob \citep{shi2017world} and Miniwob++ \citep{liu2018reinforcement} environments, but relied on obtaining expert demonstrations for each website, which cannot scale effectively to cover the large variety of real-world websites, and cannot adapt to changing websites. Further, these methods failed to solve complex web navigation tasks such as flight booking or social media interaction \citep{gur2018learning}. 

\citet{gur2018learning} take a step farther by training an RL agent to solve complex web navigation tasks using a scheduled curriculum. The curriculum linearly increases a parameter $p$, in which $1-p$ controls the number of web elements that are solved by querying an oracle policy, which is obtained via expert data. 
This work differs in several ways. First, we do not rely on any expert demonstrations to augment sparse rewards. We use AEG to automatically learn to generate a curriculum of web navigation tasks that are tailored to the current skill level of the agent. 
Next, we make no assumption on the availability of any website while they assume websites are given \textit{a priori}. Lastly, our web navigation agents generalize to unseen environments. 

Multi-agent training can be an effective method for automatically generating a curriculum of RL tasks (e.g. ~\citet{leibo2019autocurricula,matiisen2019teacher,graves2017automated,portelas2020teacher}). For example, Asymmetric Self Play (ASP) \citep{sukhbaatar2017intrinsic} trains two agents, in which the second agent must learn to repeat the actions taken by the first, demonstrator agent.
Both agents play in the same, fixed environment. In contrast, we use a third agent to learn to generate challenging new environments. 
POET \citep{wang2019paired,wang2020enhanced}
is an AEG technique which uses a population of adversaries to generate the terrain a 2D walker agent must learn to navigate.
To create a curriculum, POET requires generating many new environments, testing all agents within each one, and discarding environments based on a manually chosen a reward threshold, which wastes a significant amount of computation. 
\citet{campero2020learning} use a teacher to propose navigation tasks; the teacher's reward is based on whether the agent takes more steps than a threshold, a hyperparmeter that is linearly increased over the course of training.

Most closely related to our work is PAIRED \citep{dennis2020emergent}, which is an AEG method for training agents with minimal regret that works by constraining the environment-generating adversary using the performance of a second agent. However, PAIRED only demonstrated results on simple gridworld environments, and did not expand to the type of complex, high-dimensional state-action space required for web navigation. We improve on PAIRED using a more flexible estimate of the regret, as well as a budget mechanism, and show that this significantly improves performance.

%% file: src/background.tex
\section{Background}

\subsection{Web Navigation Problem}

Following previous work \citep{shi2017world,gur2018learning,liu2018reinforcement}, we formulate web navigation as a sequential decision making problem where we train an agent, parameterized by a network $\pi(a_t | s_t; \Theta_i)$, that maps an input state $s_t$ to output actions $a_t$ to maximize the cumulative discounted reward, .i.e.,  $O=\sum_{t=0}^{T}{\gamma^t r_t}$ where $r_t$ is the reward at time step $t$, $\gamma$ is a discount factor, and $T$ is the length of an episode.
We use the web page and user instruction as the input state. The web page is dynamically updated at each time step, while the instruction is fixed at the beginning of an episode.
We represent web pages using Document Object Model (DOM), a tree of elements in a page, where each element is denoted by a set of (attribute, value) pairs and an array of features (such as spatial coordinates).
Instructions are given as a set of fields where each field is a (key, value) pair.
Keys are fixed for each task and values dynamically change based on user input.

Each action is represented as a tuple (element, field) that denotes acting on the element using the field as an input; i.e. typing the value of the field into the element.
Agents receive a task success reward (1.0 or -1.0) at the end of each episode, a potential-based reward when the value of an element in the page is updated, and a small penalty each timestep to encourage efficient navigation.
As an example, consider a flight booking task where the agent is given an instruction \{\texttt{"Departure Date": "Friday", Destination Airport: "Los Angeles (LAX)"}\}.
The agent first picks a field (e.g. destination airport) and finds the corresponding text box in the page; then the corresponding value (``Los Angeles (LAX)") typed in to the text box.
If this value is correct, the agent receives a positive reward of $1 /  N $ where $N$ is the number of fields.

\subsection{Protagonist Antagonist Induced Regret Environment Design (PAIRED)}
Adversarial Environment Generation (AEG) trains an adversary policy $\pi_E$ to design environments to minimize the performance of an agent's policy, $\pi_P$. Let $R^P_i = \sum_{t=1}^T\gamma^tr_t^P$ be the total reward received by the agent for trajectory $i$. In minimax AEG, the objective for the adversary is simply: $-R^P$. Thus, minimax adversaries are incentivized to create excessively difficult or impossible environments, which may not enable the agent to learn. Instead, PAIRED \citep{dennis2020emergent} trains the adversary to maximize the agent's \textit{regret}, which is defined as the difference between the agent's return and the return of the optimal policy, $R^*-R^P$. When the reward function includes an incentive to complete the task more efficiently (which is true in our case), the regret will be highest for easy tasks which could be completed in a few steps by the optimal policy, but which the current policy fails to complete. Therefore, an adversary that maximizes the regret will continue to propose easier tasks until the agent begins to solve them, making regret a desirable objective for AEG.

To estimate the regret, PAIRED
introduces a third agent, the \textit{antagonist} (with policy $\pi_A$), and  constrains the adversary to only generate feasible environments which the antagonist can complete. 
When the adversary generates an environment $E$, both the protagonist and antagonist collect $M$ trajectories with returns $R^P_{1},...,R^P_{M}, R^A_{1},...,R^A_{M}$ in $E$. The regret is then estimated as:
\begin{align}
    \textsc{Regret} = \max_i R^A_i - \frac{1}{M}\sum_{m=1}^M R^P_m
    \label{eq:paired_regret}
\end{align}

As \citet{dennis2020emergent} show, if the adversary and antagonist coordinate and reach a Nash equilibrium with the protagonist, then the protagonist will have learned to minimize the regret. However, in practice gradient-based multi-agent RL has no convergence guarantees, is highly non-stationary, and will often fail to converge \citep{mazumdar2019policy,mazumdar2019finding}. If the antagonist and adversary in PAIRED fail to coordinate, then PAIRED minimizes regret with respect to the antagonist's policy. In that case, the objective in Equation \ref{eq:paired_regret} only forces the protagonist to learn to be as good as the antagonist. If the antagonist fails to improve, or reaches a local optimum, then the adversary cannot continue to train the protagonist. In Section \ref{sec:flexible} we propose an improved objective which addresses this problem. 

%% file: src/model.tex
\section{Web Environment Design}
\begin{figure}[t]
\centering
\vspace{-1.cm}
\subfloat[A \textbf{fully specified} DOM primitive where a label is created and its text is assigned.]{
  \includegraphics[width=0.28\linewidth]{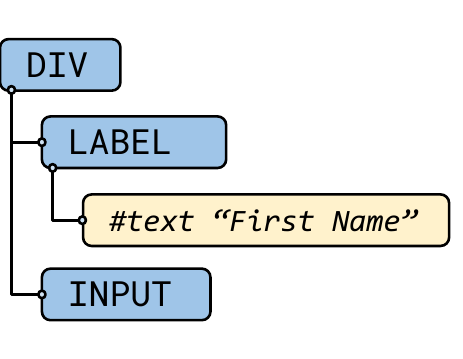}
  \label{fig:underspecified0}
}
\hspace{0.15in}
\subfloat[An \textbf{underspecified} DOM tree template. The text box is always included, its text and label element are variables.]{
  \includegraphics[width=0.28\linewidth]{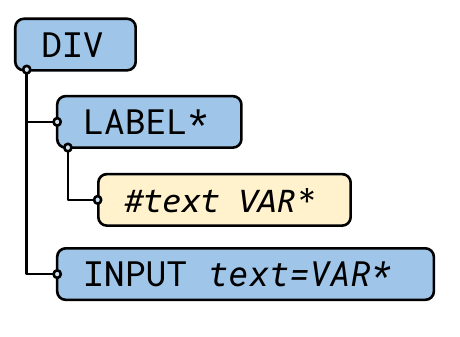}
  \label{fig:underspecified1}
}
\hspace{0.15in}
\subfloat[A \textbf{fully specified} DOM primitive where only the inner text within the text box is assigned.]{
  \includegraphics[width=0.28\linewidth]{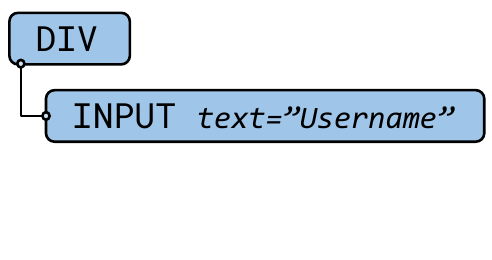}
  \label{fig:underspecified2}
}
\caption{An example underspecified DOM tree template (b) and its instantiations (a,c) with different values. (*) indicates a variable; either an element or one of its attributes. (a) is used in Page 1 and (c) is used in Page 2 in Figure \ref{fig:rollout}.}
\label{fig:underspecified}
\end{figure}

We start with an empty website that is gradually populated by new pages and links between them.
Given that we represent pages by their DOM, we focus on creating DOM trees and assume links between pages are implicitly defined by events attached to certain elements.

\begin{figure}
    \centering
    \includegraphics[scale=0.6]{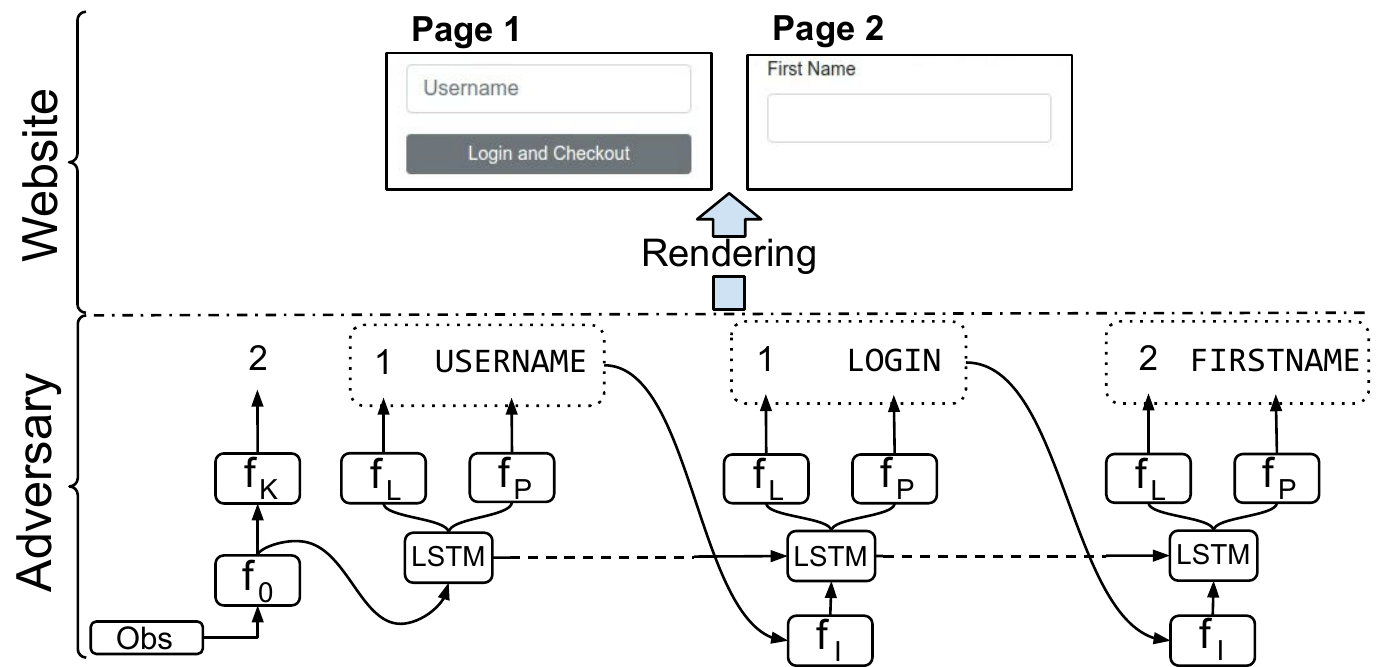}
    \caption{\small A sample rollout of the adversary for compositional environment generation for web navigation problem. An initial observation (Obs) is given at the beginning of the rollout. $f_0$, $f_K$, $f_L$, $f_P$, and $f_I$ denote networks for encoding initial observation, generating number of pages, page indices,\protect\footnotemark primitives, and encoding LSTM inputs, respectively.}
    \label{fig:rollout}
    \vspace{-0.4cm}
\end{figure}

While the most general approach to designing DOM trees would be combining a set of arbitrary elements in a bottom-up approach, this would generate a large number of malformed websites that are semantically incoherent.
Consider the second page in Figure \ref{fig:rollout} where there is a text box and a label on the top that says ``First Name''.
Now, if we have had inserted the label on top of the `Username' text box in the first page, the website would become malformed as it is ambiguous if the text box refers to `username' or `first name'.

As a result, we formulate the website design as combining a set of \textit{primitive DOM sub-trees} that are general enough to create complex websites but can be combined safely in a tree structure.
We first create a set of \textit{underspecified DOM tree templates} where certain elements and attributes are replaced with variables.
By assigning values to variables in a template, a fully specified DOM tree primitive is generated that can be combined with other primitives to create a new web page.
The order in which the primitives are combined also defines how the web page will be rendered as well.

Figure \ref{fig:underspecified} illustrates an example underspecified DOM tree template and its instantiations with different variable assignments.
We create an input template (Figure \ref{fig:underspecified1}) as a variable label and text box with a common parent.
In Figure \ref{fig:underspecified0}, we pick the label element and assign a value to its text attribute while in Figure \ref{fig:underspecified2}, we assign a value to the inner text of the text box and ignore the label element.

\footnotetext {For simplicity of illustration, we show an example generation process where primitives are generated in an increasing order of the page indices; however, in our formulation (see Section~\ref{sec:adversary} for details), the page indices corresponding to consecutive LSTM timesteps do not necessarily increase monotonically.}

\subsection{Website Design Primitives}
We introduce a new framework called gMiniWoB for automated website generation, 
which implements 40 different design primitives from 11 different underspecified DOM templates.
These primitives are widely used across the web and include `navigation bars', `product carousels', `item decks', `web forms', `item carts', `dropdowns', etc.
Every primitive includes at least one actionable element that changes the DOM structure when the agent interacts with it.
Each primitive is classified into 2 different categories based on their use in the reward computation: (i) \textbf{Active primitives (used)}, and (ii) \textbf{Passive primitives (not used)}.
$26$ of the $40$ different primitives are active primitives and the rest are passive.
When a new active primitive is added to a web page, it automatically also grows the instruction to accommodate the corresponding field.
For example, adding `First Name' text box in Figure \ref{fig:underspecified2} also adds a new field with \textit{``firstname''} key into user instruction.
This makes active primitives more complicated to learn than passive primitives, which mostly serve as noise. However, real websites contain many distracting elements (passive primitives), so it is important for agents to learn to ignore them. Appendix \ref{ap:design_primitives} details all the design primitives used, and Appendix \ref{ap:test} shows the websites in the testset.

\subsection{Adversary Architecture}
\label{sec:adversary}
We propose an autoregressive adversary policy for the compositional environment generation problem where the goal is to place a set of design primitives to a set of locations.
We parametrize the adversary with a policy $\pi_E(a^A | o^A)$ such that 
\begin{align}
    \pi_E(a^A | o^A) &= \pi(k | K)\prod_{i=0}^N{\pi(a_i|a_{0\cdots i-1}, b_{0 \cdots i-1}, k) \pi(b_i|a_{0\cdots i-1}, b_{0 \cdots i-1}, k)}
\end{align}
where $N$ is an upper limit on the number of outputs, $K$ is an upper limit on the number of locations, $a_i$ is a design primitive, $b_i$ is a location index, and $o^A$ is an initial observation.
The adversary first samples the number of locations $k$ from a parametrized Categorical distribution $Cat(0, K)$.
Conditioned on $o^A$, it executes an autoregressive model to generate a set of primitives and their corresponding locations within $[0,\cdots,k]$.

We sample $o^A$ from the standard normal distribution, similar to generative adversarial networks (GAN), to allow the adversary to diversify its design distribution.
This observation is encoded with a feed forward network $h_0=f_0(o^A)$ and $h_0$ is passed to another network $f_K$ that outputs a distribution over number of empty pages.
The same hidden state $h_0$ is passed to an LSTM network as the initial input vector and output of the LSTM is used by two independent networks $f_P$ and $f_L$ to (i) learn a distribution over design primitives and (ii) learn a distribution over locations, respectively.
We sample a primitive and a location from these distributions and they are encoded by another network $f_I$ into a hidden state which is used as the input to the LSTM at the next step.
After running for $N$ steps, sampled design actions are sent to a renderer module which generates the environment.

For the web navigation problem, $K$ denotes the number of pages in the website, locations ($b_i$) denote pages, and primitives ($a_i$) denote DOM tree primitives.
We illustrate a sample rollout of the adversary for web environment generation in Figure \ref{fig:rollout}.
We also augment the primitive design actions with a special \texttt{SKIP} action that does nothing when executed by the renderer.
This allows the adversary to control the number of primitives added.

\subsection{Flexible PAIRED}
\label{sec:flexible}
We use flexible antagonist selection to improve on the regret objective of Eq. \ref{eq:paired_regret}.
We initialize two agents $A$ and $P$. At each iteration, the adversary designs a new website and each agent collects a trajectory with return $R$ by navigating the website and the regret is:
\begin{equation}
    \textsc{REGRET} = \max \{R^A, R^P\} - 0.5 * (R^A+R^P) \label{eq:flexible_regret}
\end{equation}

This objective does not make a distinction between antagonist and protagonist agents, and instead annotates the best performing agent as the antagonist.
As long as any agent has a higher performance than the other agent, the objective will continue to improve the weakest agent.
During that time, the other agent in the policy continues learning, and therefore provide a stronger maximum performance against which we measure the regret. The Flexible PAIRED algorithm we propose is shown below. Using policy gradient updates, we train each agent in the population to optimize environmental reward, and the adversary to maximize the regret as computed in Eq. \ref{eq:flexible_regret}.

\begin{algorithm}[H]
% \SetAlgoLined
\begin{algorithmic}[1]
 \State \textbf{Input:}{$A,P$: Initialize two agents independently}
 \State $W$ $\longleftarrow$ Run the adversary $\pi_E$ to generate a new website
 \State $R^{A}$, $R^{P}$ $\longleftarrow$ Run agent $A$ and $P$ in the environment $W$ and collect rewards
 \State REGRET $\longleftarrow$ Compute regret as in Eq. \ref{eq:flexible_regret}
 \State Update adversary parameters using REGRET as the reward
 \State Update parameters of $A$ and $P$ using $R^{A}$ and $R^{P}$, respectively
\end{algorithmic}
 \caption{One step training of flexible PAIRED}
\end{algorithm}

\subsection{Budget Enforcing on Adversary}
Consider the following scenario where agents are placed on the home page of a shopping website where there are many possible elements, but only a single button that takes them to their account page.
During exploration, agents mostly collect negative rewards for taking incorrect actions, bounded to a very narrow interval (as there is only a single optimal action).
In this case, the regret is very small and non-informative, which hinders the adversary's ability to design environments at an appropriate difficulty for agents to learn.
This is true even with the proposed flexible regret objective.

To mitigate this problem, we use a budget enforcing objective in addition to the regret that binds the adversary's design budget to the performance of the best agent.
We approximate the effective budget of the adversary as the expected number of non-SKIP actions over $N$ time steps and update this budget according to whether the agents are learning.
More formally, we use the following minimization objective for budget enforcing that is added to the PAIRED objective:
\begin{equation}
    \mathcal{O}_{budget} = R^A * \sum_{i=1}^N{\log  \pi(a_i=\text{SKIP}|a_{0\cdots i-1}, b_{0,\cdots, i-1})}
\end{equation}
where $R^A$ is the reward of the antagonist (or the best-performing) agent.
This objective encourages the adversary to use less budget (more SKIP actions) when the agents are not yet learning (i.e.,  $R^A$ is negative or low); it encourages the adversary to use more budget (less SKIP actions) when the \webnav\ agents are collecting positive rewards in the environment.

%% file: src/experiments.tex
\section{Experiments and Methods}
We evaluate our models on a variety of web environments implemented in MiniWoB framework \citep{shi2017world,liu2018reinforcement}.
We implemented several challenging websites with varying difficulty levels using the same set of design primitives.
These environments include `Login', `Enter Address', `Flight Booking', `Enter Payment', and `Shopping' websites, where the agents need to enter text or select information in the website while navigating between pages.
Each environment comes with 4 different difficulty levels by gradually adding more primitives to websites.
These environments are never explicitly presented to agents during training, so performance in them measures how well agents can generalize to unseen websites at test time.

\textbf{Agent architecture:} Following \citet{gur2018learning}, we utilize an LSTM based DOM tree encoder and a feed forward network to encode profile fields.
The \webnav\ agent policy outputs a joint distribution over elements and fields by measuring pairwise similarities between element encodings and profile fields.
We compute the state-value by using the marginal distribution of elements as attention weights over element encodings and passing the context vector through a FF network.
Web navigation agents are trained with an actor-critic algorithm \citep{liu2018reinforcement}.
We train the LSTM-based adversary network using \WEDP\ and \WEDPb\ with policy gradient.

\textbf{Baselines:} We benchmark PAIRED, \WEDP, and \WEDPb\ against two additional baselines. First, a Domain Randomization (DR) agent, which we implement using a similar approach as \citet{dennis2020emergent}.
We first sample the number of empty pages $k$ from a uniform distribution $U[0,K]$.
Next, we randomly sample a primitive (including SKIP), and a page from $U[0,k]$ for $N$ steps.
Second, a Curriculum Learning (CL) approach, which adapts the scheduled curriculum idea of \citet{gur2018learning} to zero-shot environment generation where we are not given a specific website but a set of design primitives.
We randomly sample each primitive w.r.t. a probability $p$ where $p$ is initialized with a small number and scheduled to reach $1.0$ during training.

\section{Results}
\begin{figure}[tb]
\small
\centering
\vspace{-.7cm}
    \subfloat[Login]{
      \includegraphics[width=0.28\linewidth]{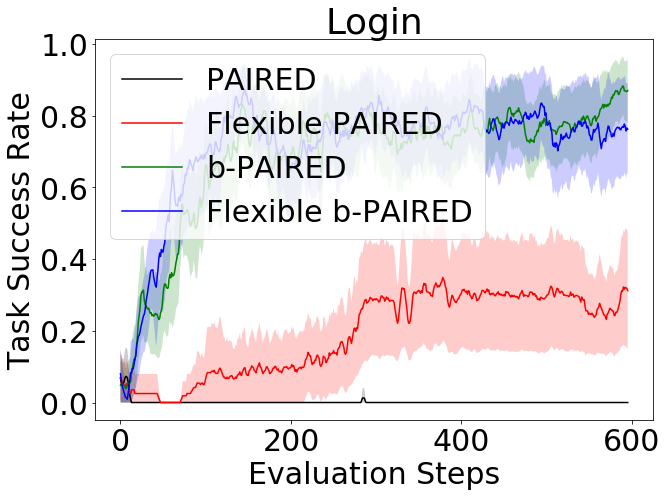}
    }%
    \subfloat[Address]{
      \includegraphics[width=0.28\linewidth]{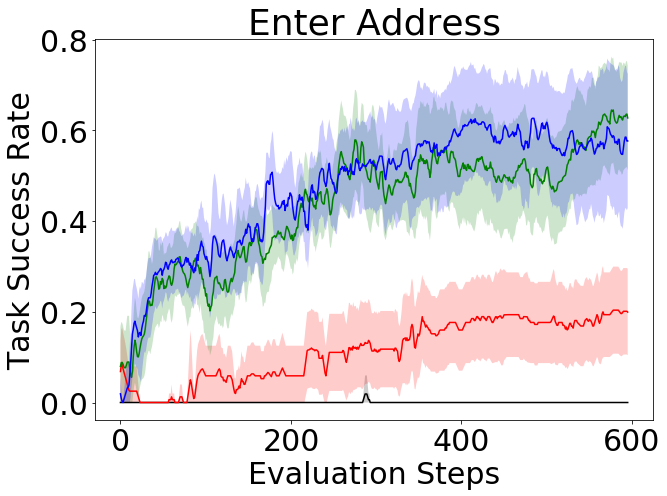}
    }%
    \subfloat[Payment]{
      \includegraphics[width=0.28\linewidth]{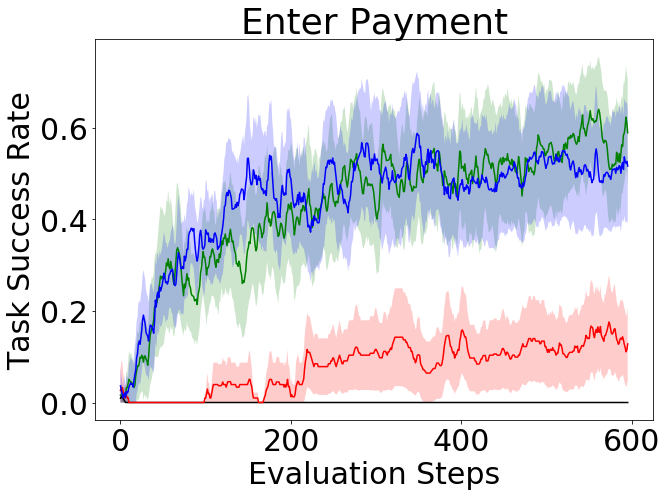}
    }\\%
    \vspace*{0.05in}
    \subfloat[Shopping]{
      \includegraphics[width=0.28\linewidth]{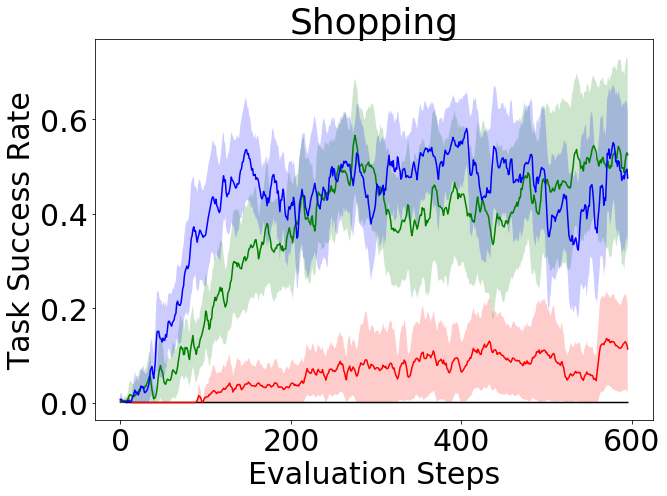}
    }%
    \subfloat[Flight booking]{
      \includegraphics[width=0.28\linewidth]{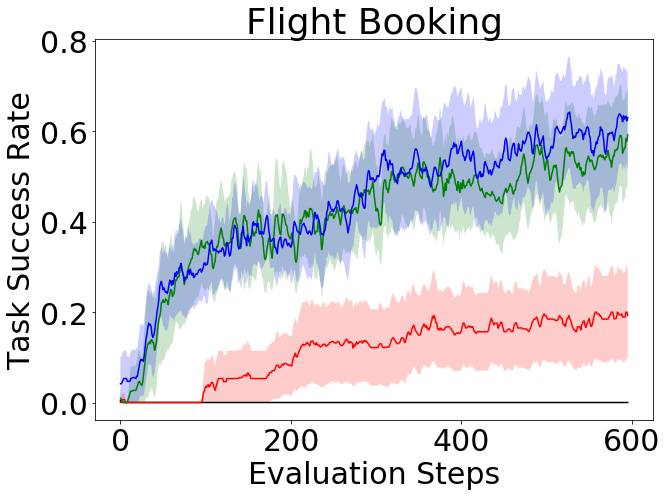}
    }%
    \subfloat[Primitives]{\includegraphics[width=0.28\linewidth]{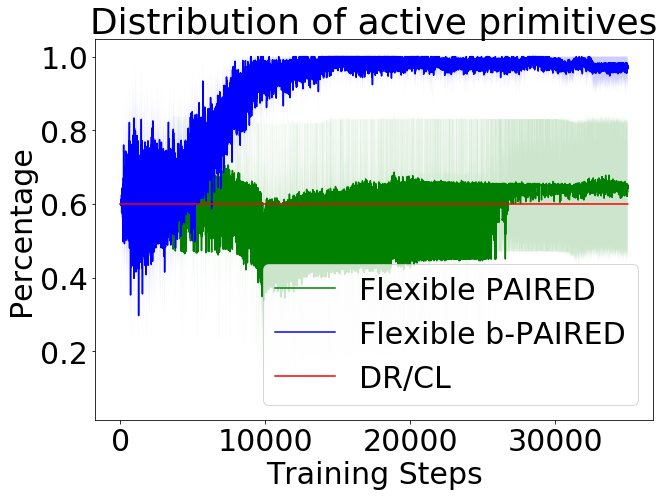}\label{fig:primitivedistribution}}
    \caption{\small Comparison of PAIRED \citep{dennis2020emergent} and Flexible PAIRED with and without budget enforcing; averaged over 4 difficulty levels. (f): Percentage of active primitives over training steps.}
    \label{fig:combinedpbt}
\end{figure}

\begin{figure}[tb]
\small
    \subfloat[Difficulty level 1]{
      \includegraphics[width=0.245\linewidth]{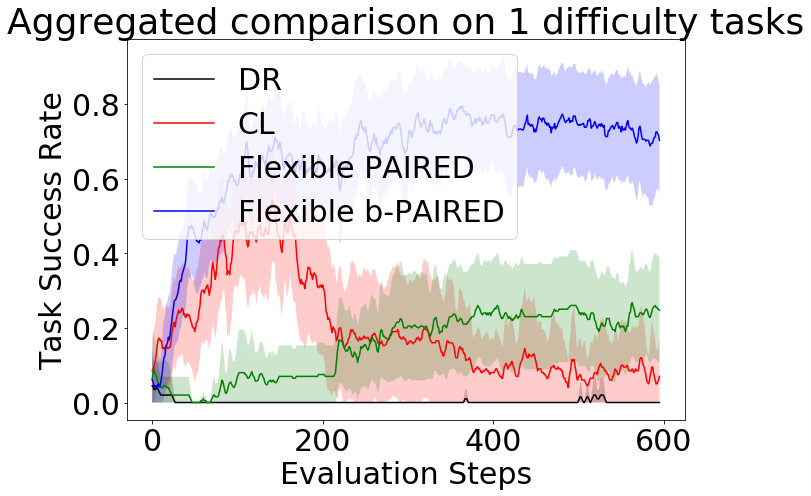}
    }%
    \subfloat[Difficulty level 2]{
      \includegraphics[width=0.245\linewidth]{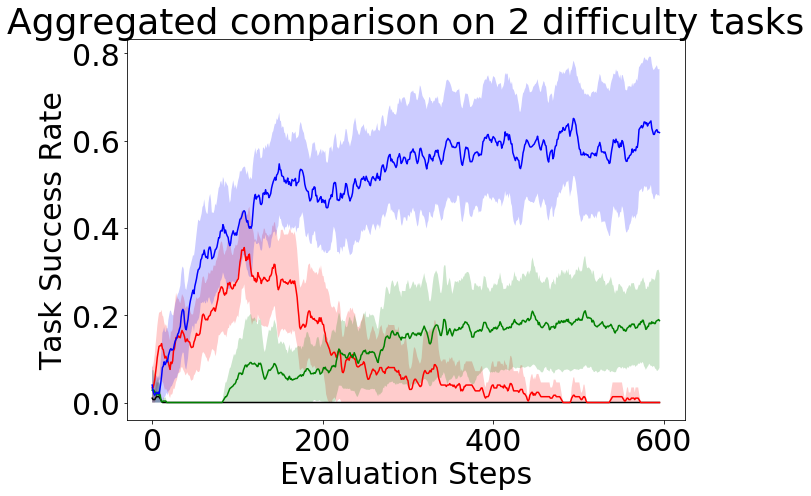}
    }%
    \subfloat[Difficulty level 3]{
      \includegraphics[width=0.245\linewidth]{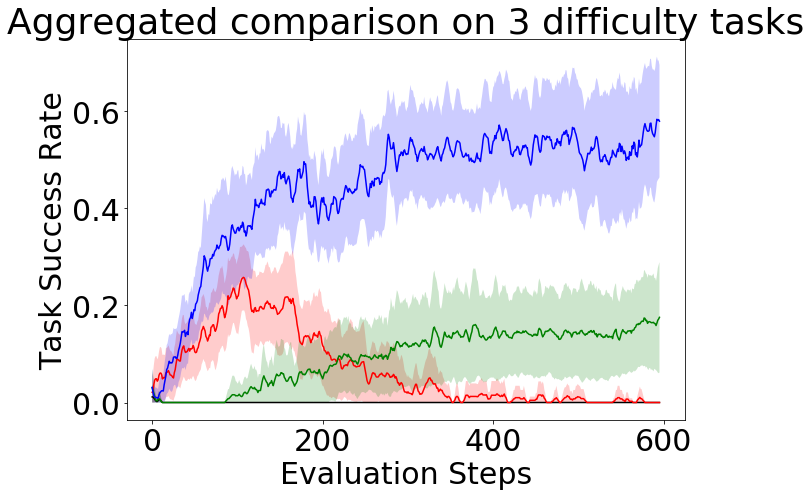}
    }%
    \subfloat[Difficulty level 4]{
      \includegraphics[width=0.245\linewidth]{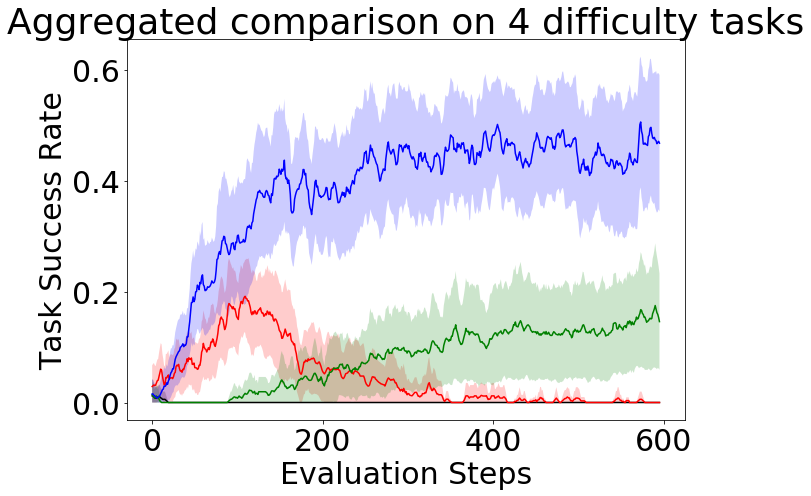}
    }%
    \caption{\small Aggregated task success rate comparison of \WEDPb\ and baseline models on test environments with increasing difficulty levels. See Appendix \ref{ap:detailed-results} for detailed results.}
    \label{fig:tasksuccessaggregated}
    \vspace{-.7cm}
\end{figure}
We first compare the original PAIRED algorithm (which used separate antagonist and protagonist agents) to the proposed \WEDP\ algorithm that annotates the best performing agent as the antagonist.
\WEDP\ considerably improves upon PAIRED, which fails to learn in this environment (Figure \ref{fig:combinedpbt}).
One reason is that when agents are separate and have very similar rewards, especially early during training, the regret becomes very small. This uninformative signal makes it difficult for the adversary to learn.
On the other hand, Flexible PAIRED computes a consistently positive regret signal, which more clearly indicates to the adversary which environments are challenging, but still feasible. 
The further ablation studies show that adding budget improves performance for both flexible, and original PAIRED method. 

\textbf{Comparison on test environments:}
We  evaluate the performance of the proposed models and baselines on task success rate computed across test environments with different difficulty levels.
\WEDPb\ outperforms \WEDP\, indicating the budget objective significantly improves performance (Figure \ref{fig:tasksuccessaggregated}). Further, both techniques significantly outperform the baseline models on all tasks, with \WEDPb\ effectively reaching more than 80\% task success on difficulty 1 tasks.
Even as the complexity of the environments continues to increase (see Section \ref{sec:complexity}), \WEDPb\ agents still perform consistently well without degrading performance.
While CL outperforms \WEDP\ early in the training, its performance drops significantly due to ignoring agents' skill level, and making environments that are too challenging for agents to complete.
We also observe that \WEDPb\ learns faster than \WEDP\ on all environments as \WEDPb\ reacts to agents' performance faster than \WEDP{} (see Appendix \ref{ap:detailed-results}).

\label{sec:complexity}
\textbf{Environments complexity:} While agent performance improves over time, we would like to know if they are presented with more challenging environments over training.
We estimate the percentage of active primitives generated as a measure of environment complexity.
Learning a web page with more passive primitives is a relatively easier task than a page with more active primitives, because passive primitives either add noise and should ignored by the agents, or are used by agents only to navigate to another page.
On the other hand, if there are more active primitives, not only will the size of the DOM tree increase but the number of profile fields will increase, making the matching between elements and profile more challenging.
\WEDPb\ starts around 60\% random selection of primitives, and gradually generates more active primitives (Figure \ref{fig:primitivedistribution}).
Although presented with more active primitives by \WEDPb, agents are still able to improve thanks to \WEDPb's ability to accurately tune the difficulty of the environments according to agents' skill. We also observe that the distribution of the primitives shifts later in the training to more complex and relevant primitives (see Appendix \ref{ap:primitives}). 

%% file: src/conclusion.tex
\vspace*{-0.1in}
\section{Conclusion}
\vspace*{-0.1in}
This work presents a novel technique for Adversarial Environment Generation (AEG), which we show improves significantly over prior work. In addition, we apply AEG to the problem of web navigation, and provide an open-source environment that enables learning to design complex websites out of a set of compositional primitives. Our \WEDPb\ method is able to generate a curriculum of increasingly complicated websites, and successfully trains agents which can navigate challenging, high-dimensional websites. 

%% file: src/appendix.tex
\newpage
\section{Appendix}

\subsection{Distribution of Primitives During Training}
\label{ap:primitives}
\begin{figure}
    \subfloat[Early]{
      \includegraphics[width=1.0\linewidth]{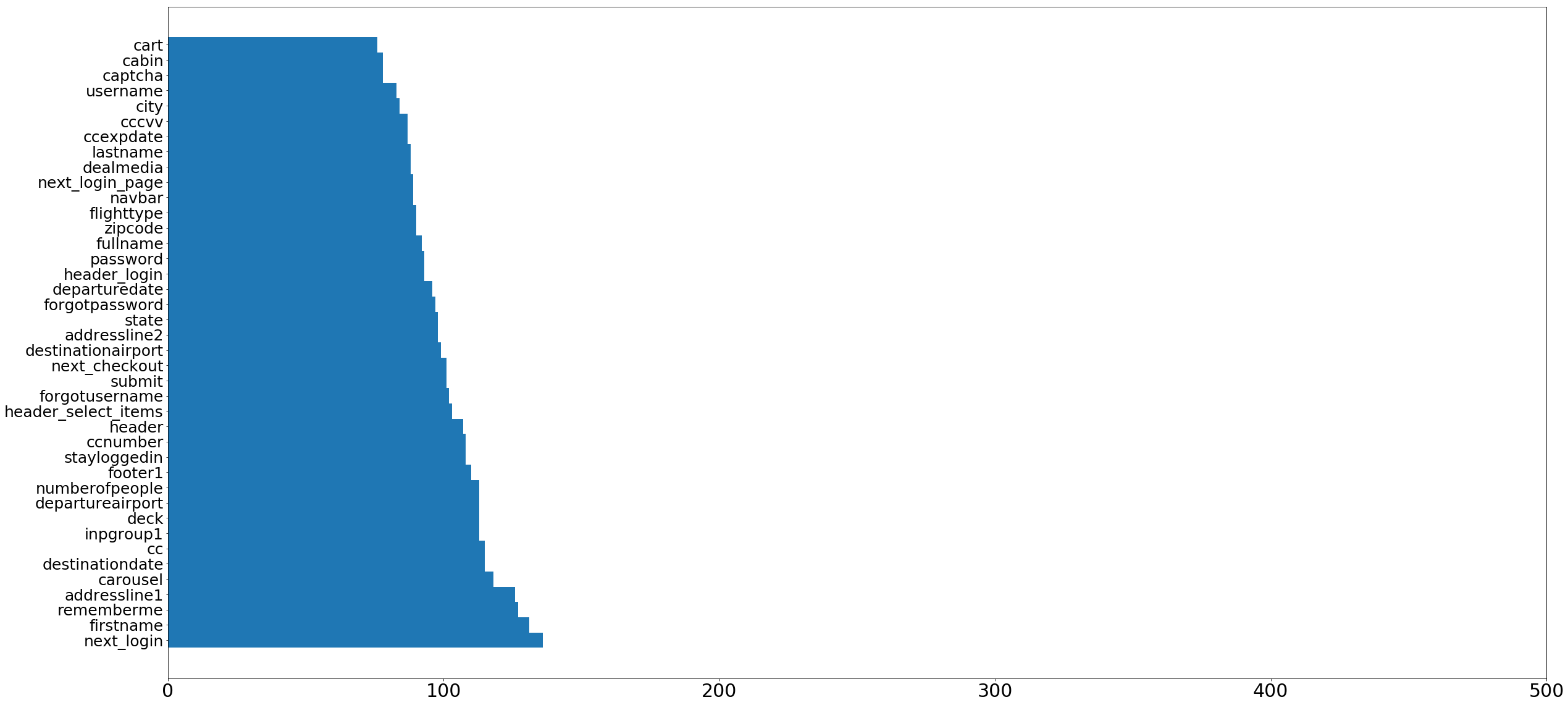}
    }\qquad
    \subfloat[Middle]{
      \includegraphics[width=1.0\linewidth]{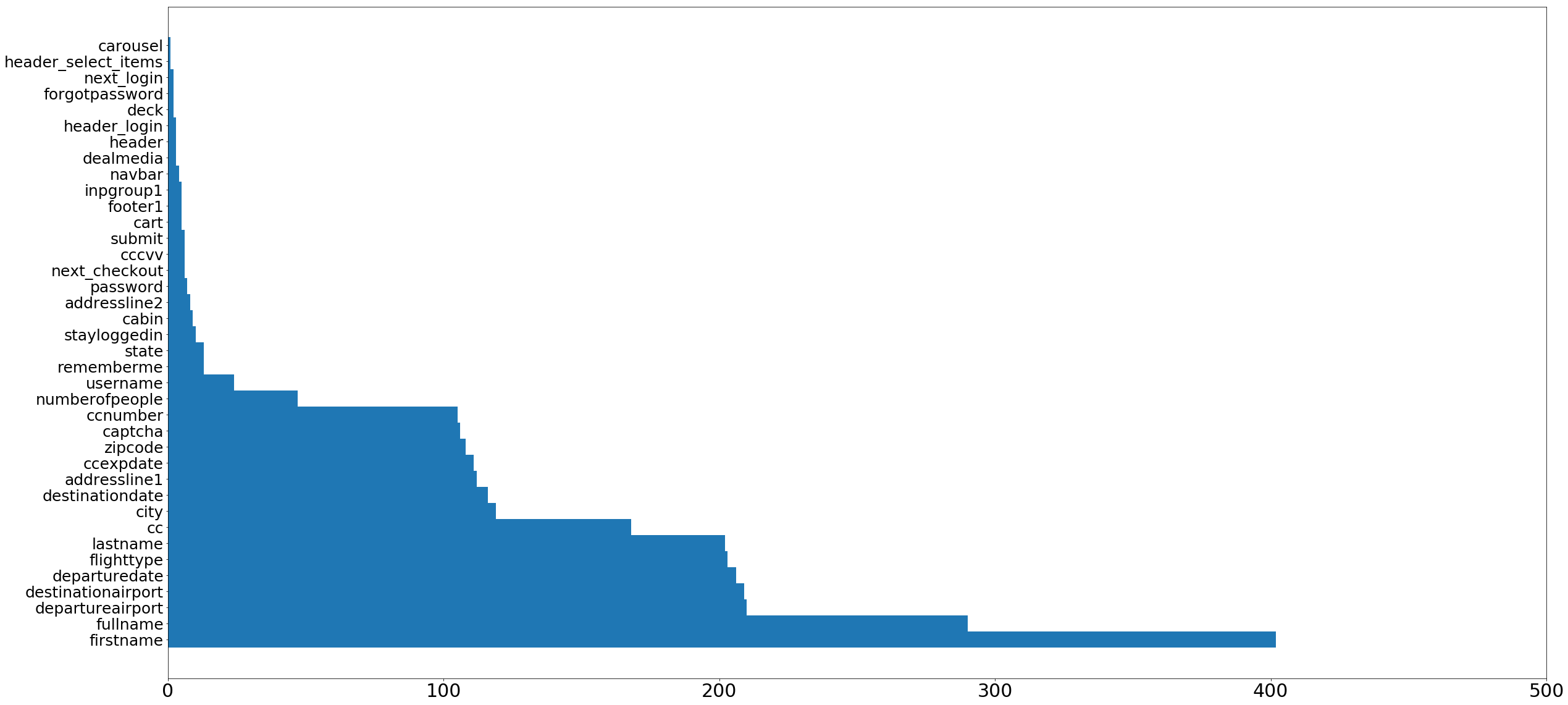}
    }\qquad
    \subfloat[Late]{
      \includegraphics[width=1.0\linewidth]{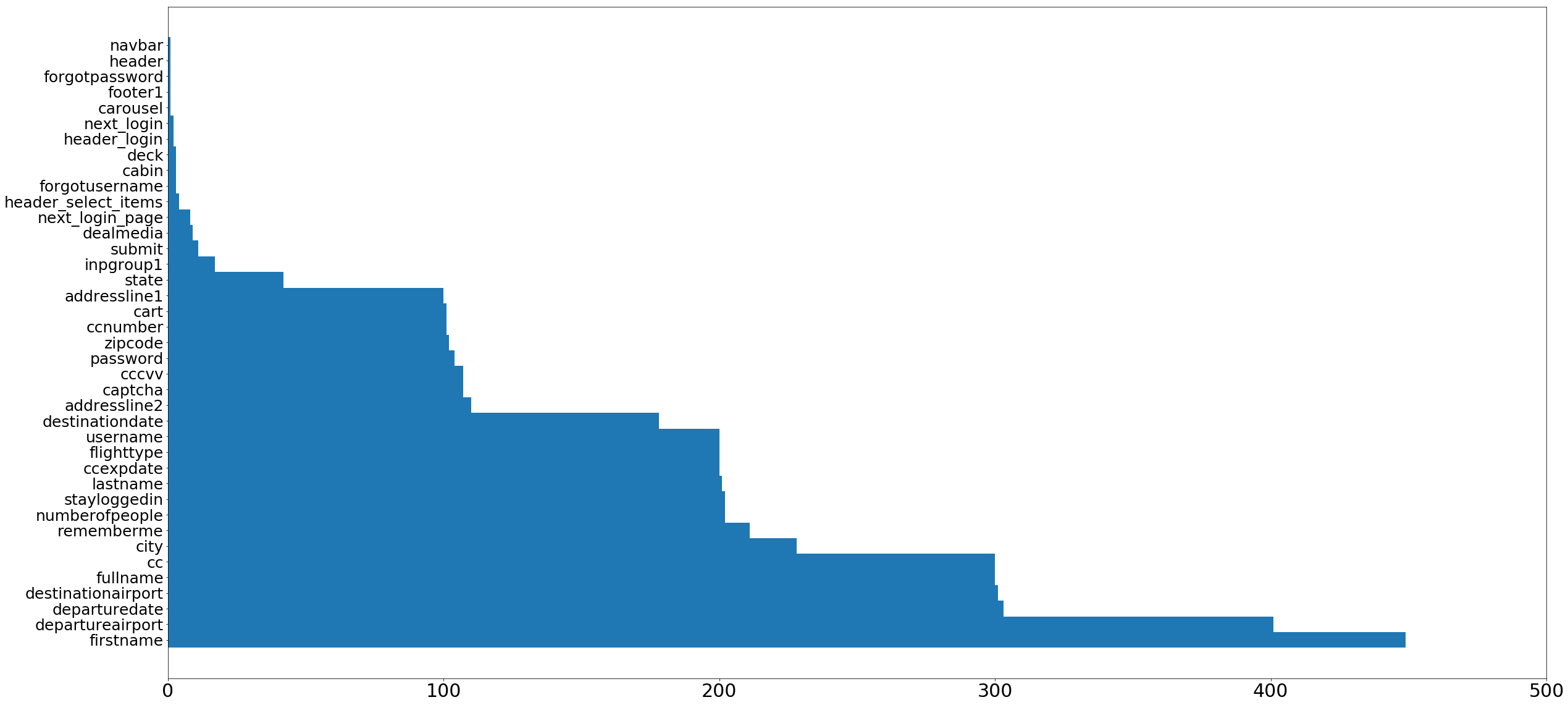}
      \label{fig:primitivehist3}
    }
    \caption{Histograms of primitives from early, middle, and late snapshots of the training.}
    \label{fig:primitivehist}
\end{figure}
During training, the distribution of primitives become more skewed towards active primitives (as shown in Figure \ref{fig:primitivedistribution}), but as the environments get more challenging, new primitives are slowly introduced as well (Figure \ref{fig:primitivehist}).
What we observe from the histograms in Figure \ref{fig:primitivehist} is that new primitives are slowly introduced between middle and late snapshots while the ranking of the primitives is also slightly changed.
For example, the adversary prefers 'departureairport' primitive more than 'fullname' primitive in the late snapshot of the training.

\subsection{Detailed Results on Test Environments}
\label{ap:detailed-results}
We detail the aggregated results in Figure \ref{fig:tasksuccessaggregated} and present performance of agents across tasks and difficulty levels (Figure \ref{fig:tasksuccess}).
On the easiest level of tasks, CL achieves slightly lower performance than \WEDPb\ early in the training while as the task difficulty increases, the gap becomes more apparent.
We observe that the primitive distribution in Figure \ref{fig:primitivehist3} and task success rate results are consistent in which late in the training, the adversary focuses more on the 'Flight Booking' related primitives and its performance still strongly increases.
\begin{table}
    \small
    \begin{tabular}{c c c c c}
    & difficulty level = 1 & difficulty level = 2 & difficulty level = 3 & difficulty level = 4
    \\ 
    \raisebox{2.5\normalbaselineskip}[0pt][0pt]{\rotatebox[origin=c]{90}{\small Login}} & \includegraphics[width=0.22\linewidth]{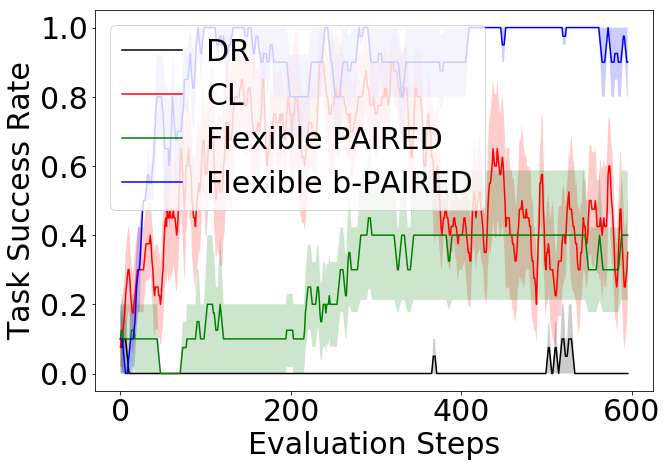}
    &\includegraphics[width=0.22\linewidth]{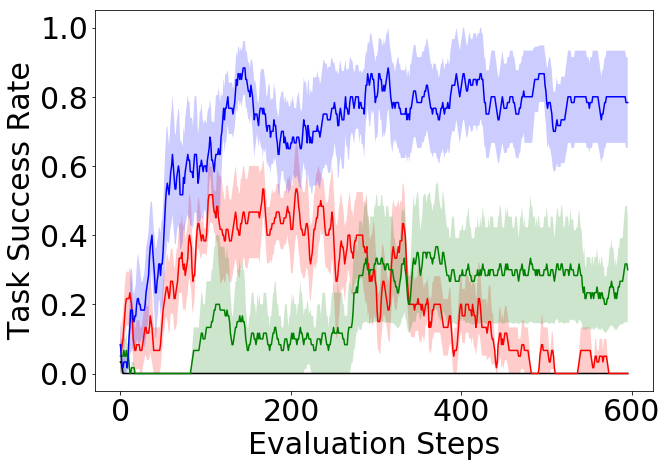}
    &\includegraphics[width=0.22\linewidth]{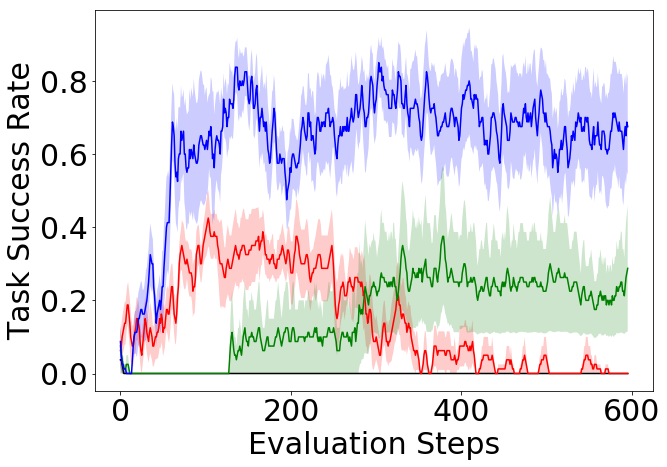}
    &\includegraphics[width=0.22\linewidth]{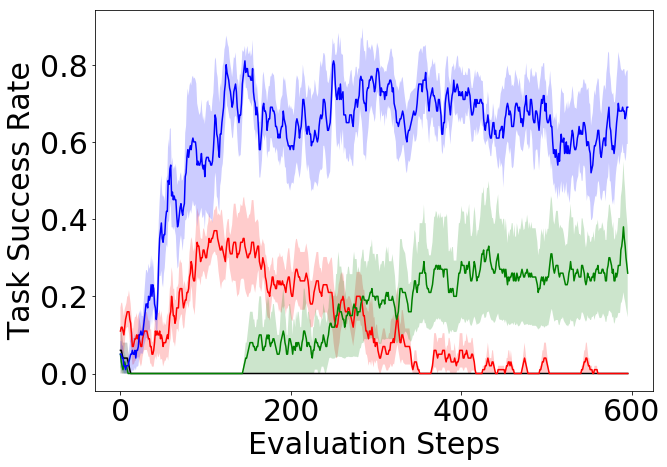}
    \\
    \raisebox{2.5\normalbaselineskip}[0pt][0pt]{\rotatebox[origin=c]{90}{\small Enter Address}} &\includegraphics[width=0.22\linewidth]{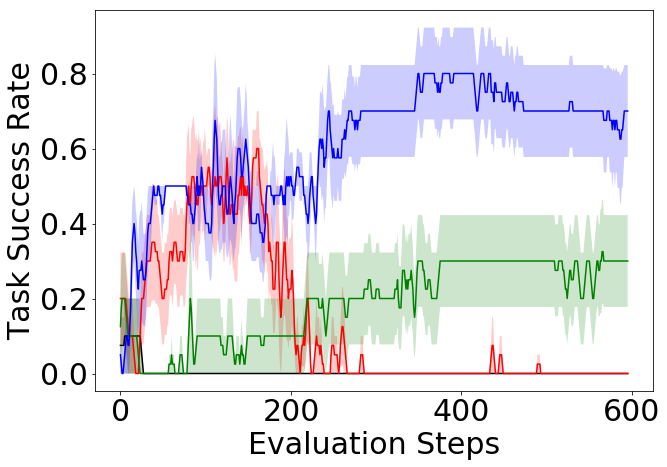}
    &\includegraphics[width=0.22\linewidth]{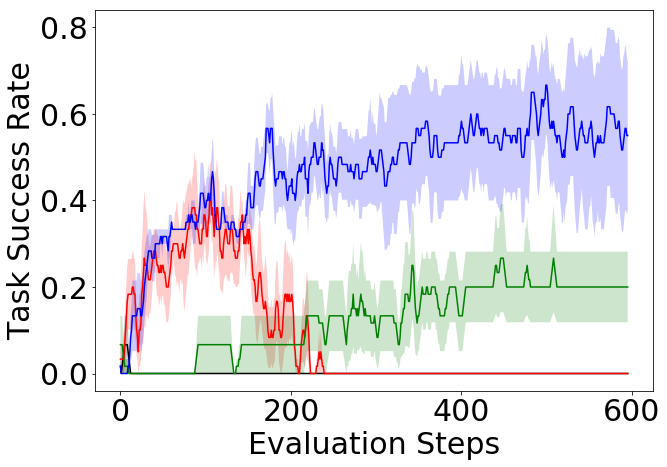}
    &\includegraphics[width=0.22\linewidth]{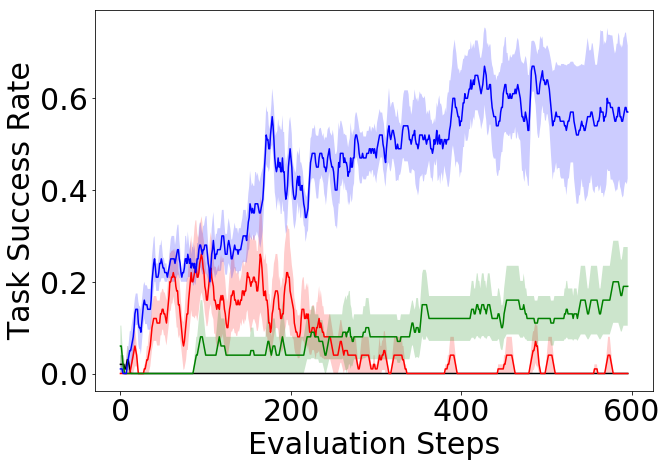}
    &\includegraphics[width=0.22\linewidth]{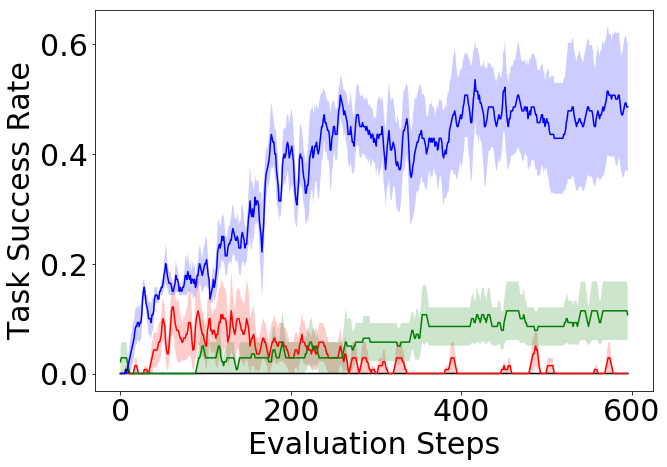}
    \\
    \raisebox{2.5\normalbaselineskip}[0pt][0pt]{\rotatebox[origin=c]{90}{\small Enter Payment}} &\includegraphics[width=0.22\linewidth]{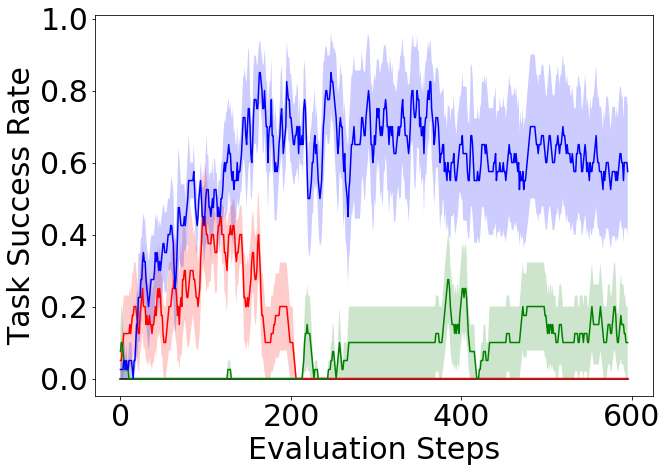}
    &\includegraphics[width=0.22\linewidth]{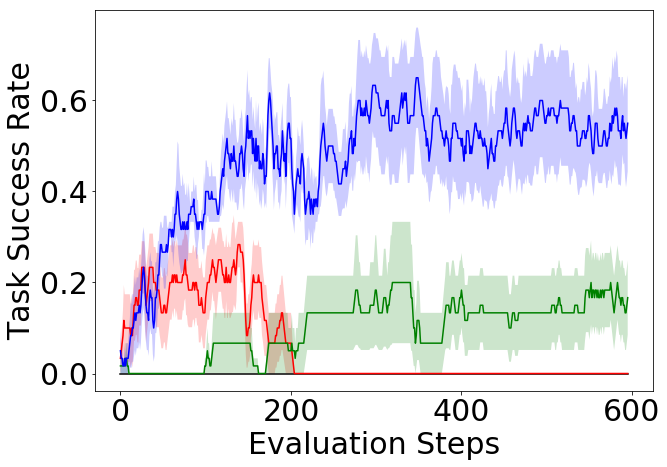}
    &\includegraphics[width=0.22\linewidth]{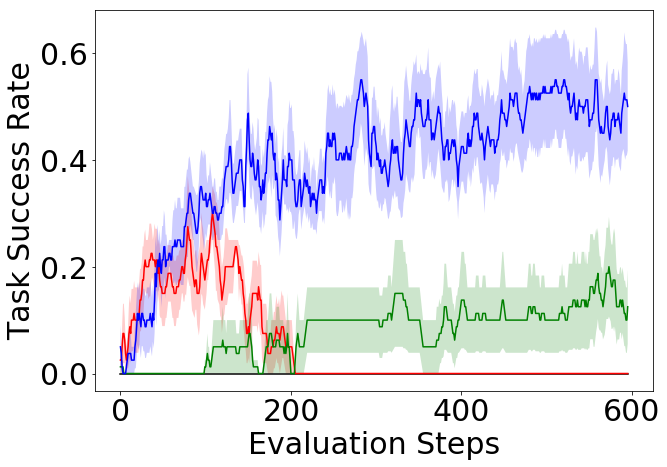}
    &\includegraphics[width=0.22\linewidth]{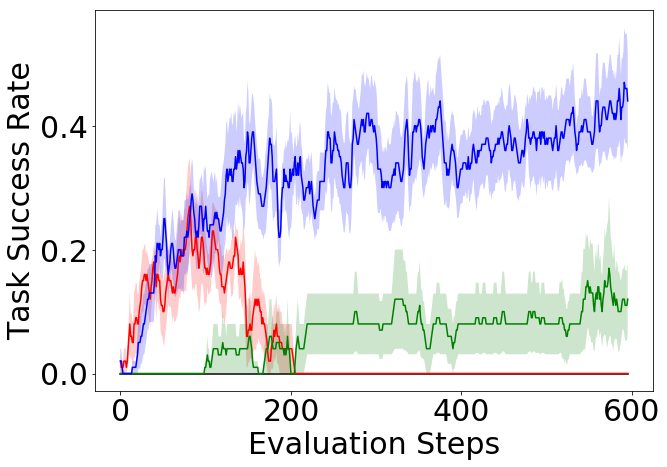}
    \\
    \raisebox{2.5\normalbaselineskip}[0pt][0pt]{\rotatebox[origin=c]{90}{\small Shopping}} &\includegraphics[width=0.22\linewidth]{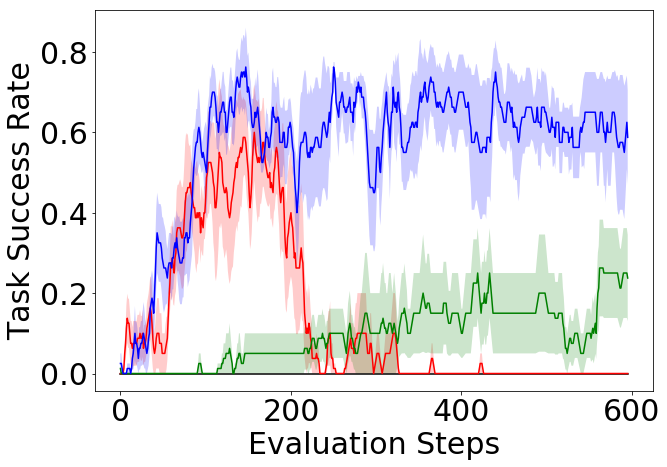}
    &\includegraphics[width=0.22\linewidth]{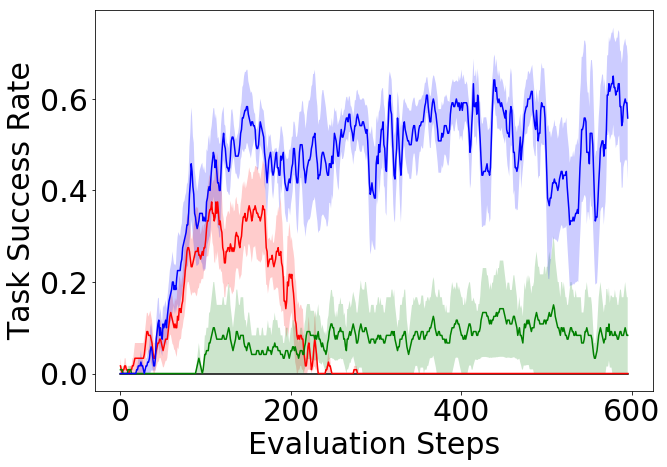}
    &\includegraphics[width=0.22\linewidth]{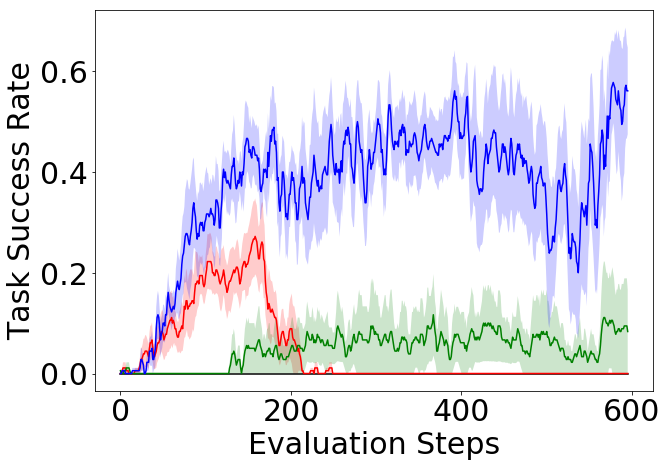}
    &\includegraphics[width=0.22\linewidth]{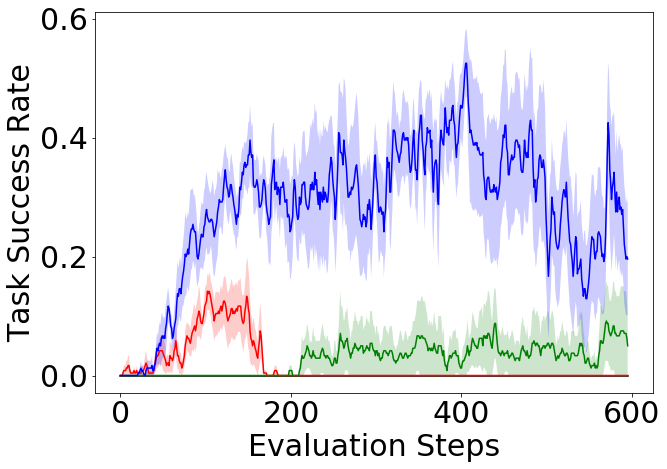}
    \\
    \raisebox{2.5\normalbaselineskip}[0pt][0pt]{\rotatebox[origin=c]{90}{\small Flight Booking}} &\includegraphics[width=0.22\linewidth]{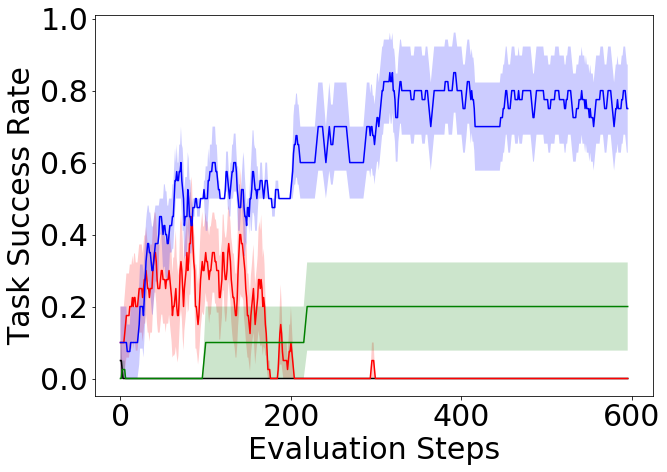}
    &\includegraphics[width=0.22\linewidth]{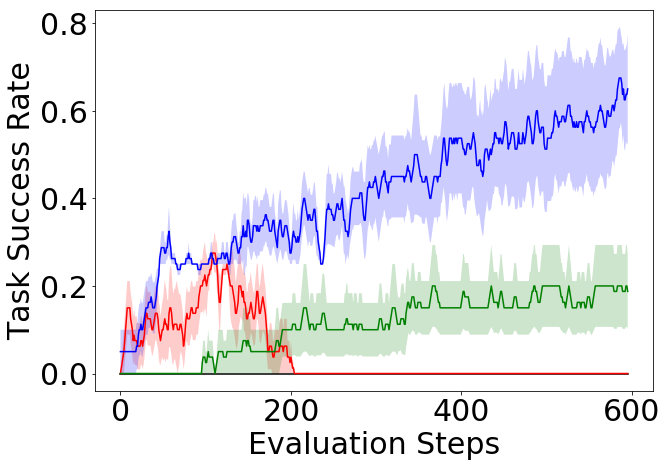}
    &\includegraphics[width=0.22\linewidth]{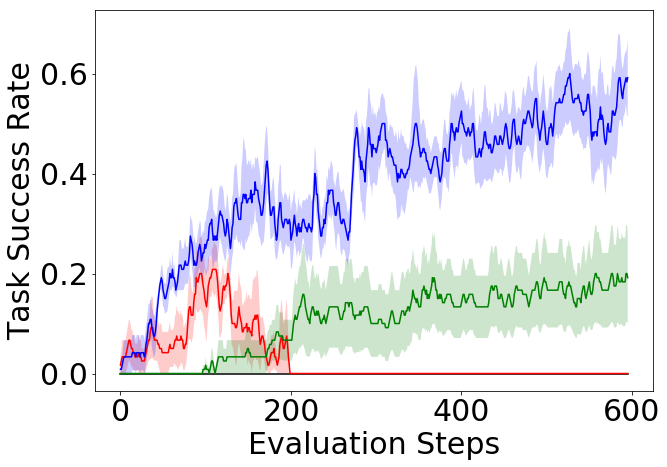}
    &\includegraphics[width=0.22\linewidth]{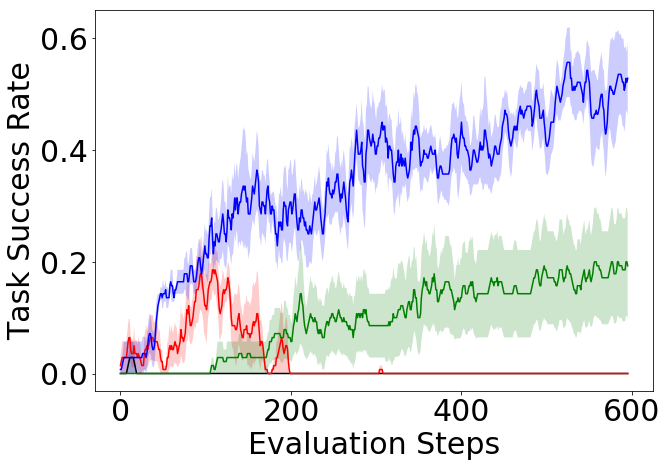}
    \\
    \end{tabular}
    \caption{Task success rate comparison of PAIRED and baseline models on test environments with increasing difficulty levels. From left to right, columns correspond to increasing difficulty. From top to bottom, rows correspond to different test environments.}
    \label{fig:tasksuccess}
\end{table}

\subsection{Web Environment Design Primitives}
\label{ap:design_primitives}
\begin{center}
\begin{tabular}{ r|c|c|p{5cm}  }
 %\hline
 \multicolumn{4}{c}{\large \textbf{Design Primitives and Their Descriptions}} \\
 \hline 
 \textbf{Design Primitive} & \textbf{Design Template} & \textbf{Active/Passive} & \textbf{Description}\\
 \hline \hline
    addressline1   & input & active & Main address information\\ \hline
    addressline2 &  input  & active & Secondary address information\\ \hline
    cabin & multi-selection & active & Multiple cabin options\\ \hline
    captcha & input & active & Captcha information\\ \hline
    carousel & carousel & passive & Items with images in a carousel with previous and next buttons \\ \hline
    cart & cart & passive & Items in a product cart with promo code information\\ \hline
    cc & multi-selection & active & Multiple credit card type options\\ \hline
    cccvv & input & active & Credit card CVV information\\ \hline
    ccexpdate & input & active & Credit card expiration date information \\ \hline
    ccnumber & input & active & Credit card number information\\ \hline
    city & input & active & City address information\\ \hline
    dealmedia & media & passive & Product media with image, label, and link\\ \hline
    deck & deck & passive & Multiple product decks with image, label, and link\\ \hline
    departureairport & input & active & Departure airport information \\ \hline
    departuredate & input & active & Departure date information\\ \hline
    destinationairport & input & active & Destination airport information \\ \hline
    destinationdate & input & active & Destination date information\\ \hline
    firstname & input & active & First name information\\ \hline
    flighttype & multi-selection & active & Multiple flight type options\\ \hline
    footer1 & footer & passive & Footer with links and information\\ \hline
    forgotpassword & link & passive & Link with forgot password context \\ \hline
    forgotusername & link & passive & Link with forgot username context\\ \hline
    fullname & input & active & First and last name information\\ \hline
    header & label & passive & Generic header\\  \hline
    header\_login & label & passive & Header for login form\\ \hline
    header\_select\_items & label & passive & Header for item selection\\ \hline
    inpgroup1 & input & passive & Generic input with default search context\\ \hline
    lastname & input & active & Last name information\\ \hline
    navbar & navigation bar & passive & A navigation bar with a menu \\ \hline
    next\_checkout & button & passive & Next button with checkout context \\ \hline
    next\_login & button & passive & Next button with login context \\ \hline
    next\_login\_page & button & passive & Next button with login context \\ \hline
    numberofpeople & multi-selection & active & Multiple number of people options\\ \hline
    password & input & active & Password information \\ \hline
    rememberme & selection & active & Checkbox with remember me context\\ \hline
    state & input & active & State information \\ \hline
    stayloggedin & selection & active & Checkbox with stay logged in context\\ \hline
    submit & button & passive & Submit button \\ \hline
    username & input & active & Username information\\ \hline
    zipcode & input & active & Zipcode information \\ \hline
\end{tabular}
\label{table:primitives}
\end{center}
In Table \ref{table:primitives}, we present the list of design primitives, corresponding templates, types, and descriptions.

\subsection{List of Test Environments}
\label{ap:test}
\begin{figure}
    \subfloat[Login]{
      \includegraphics[width=0.23\linewidth,fbox]{img/sampledesigns/logintask.png}
    }%
    \subfloat[Enter Address]{
      \includegraphics[width=0.23\linewidth,fbox]{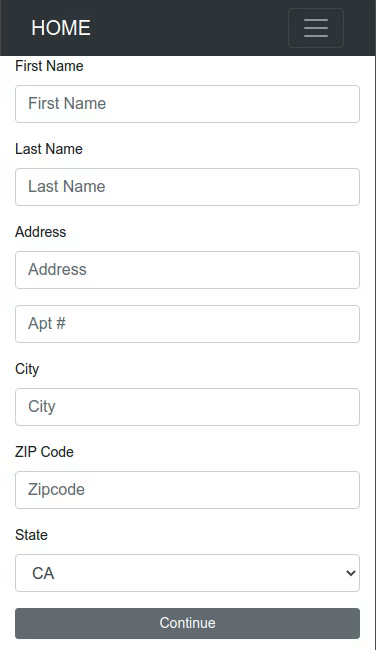}
    }%
    \subfloat[Enter Payment]{
      \includegraphics[width=0.23\linewidth,fbox]{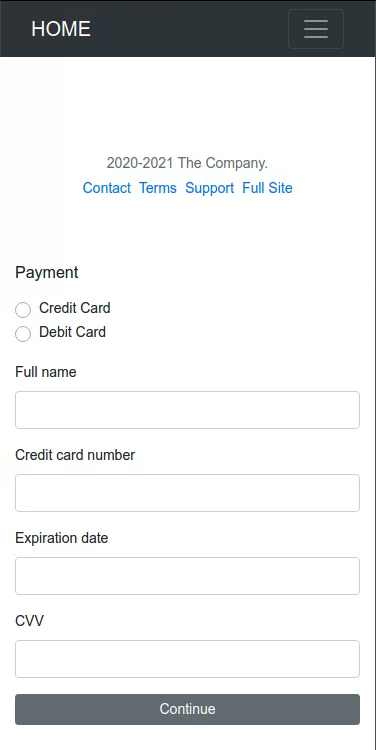}
    }%
    \subfloat[Flight Booking]{
      \includegraphics[width=0.23\linewidth,fbox]{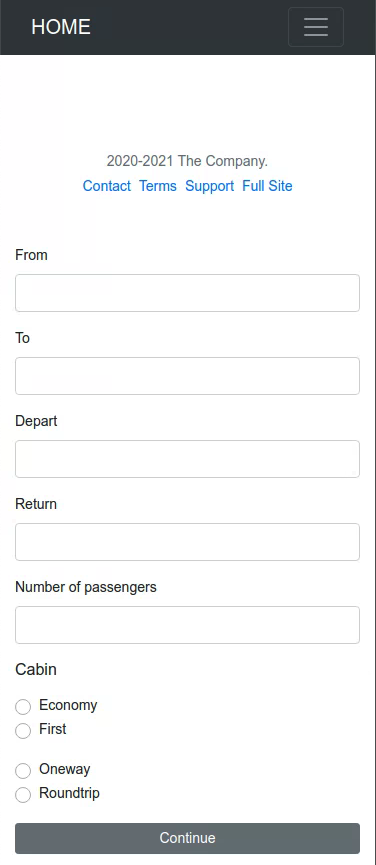}
    }%
    \caption{Screenshots of single page test environments.}
    \label{fig:testenvs}
\end{figure}

\begin{figure}
    \centering
    \subfloat[Home Page]{
      \includegraphics[width=0.23\linewidth,fbox]{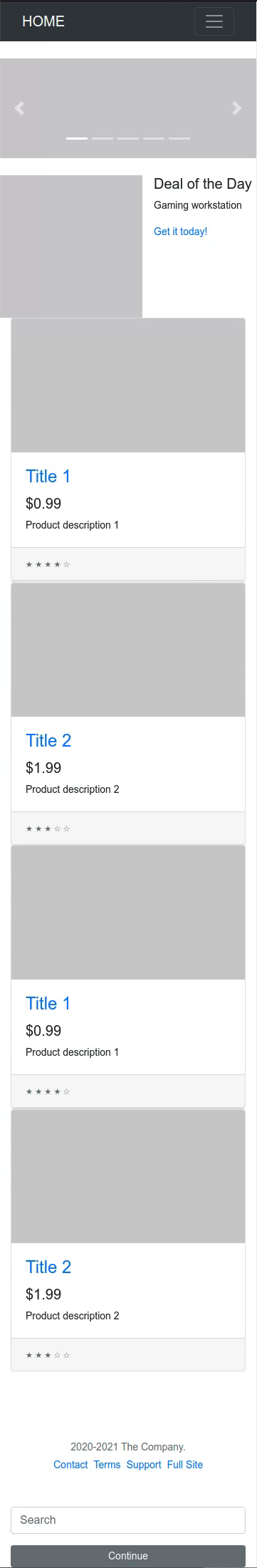}
    }%
    \subfloat[Login Page]{
      \includegraphics[width=0.23\linewidth,fbox]{img/sampledesigns/logintask.png}
    }%
    \subfloat[Address Page]{
      \includegraphics[width=0.23\linewidth,fbox]{img/sampledesigns/addresstask.png}
    }%
    \caption{Screenshots of multi-page ``Shopping'' environment. The ``Shopping'' environment is composed of a complex home page and additional ``Login'' and ``Enter Address'' pages.}
    \label{fig:shoppingtestenv}
\end{figure}

In Figure \ref{fig:testenvs} and \ref{fig:shoppingtestenv}, we present screenshots of the testing environments with the hardest difficulty levels.
While ``Login'', ``Enter Address'', ``Enter Payment'', and ``Flight Booking'' are single page environments, ``Shopping'' is a multi-page environment where an agent needs to first navigate the home page and then solve ``Login'' and ``Enter Address'' tasks.

%% file: iclr2021_conference.bbl
\begin{thebibliography}{17}
\providecommand{\natexlab}[1]{#1}
\providecommand{\url}[1]{\texttt{#1}}
\expandafter\ifx\csname urlstyle\endcsname\relax
  \providecommand{\doi}[1]{doi: #1}\else
  \providecommand{\doi}{doi: \begingroup \urlstyle{rm}\Url}\fi

\bibitem[Campero et~al.(2020)Campero, Raileanu, K{\"u}ttler, Tenenbaum,
  Rockt{\"a}schel, and Grefenstette]{campero2020learning}
Andres Campero, Roberta Raileanu, Heinrich K{\"u}ttler, Joshua~B Tenenbaum, Tim
  Rockt{\"a}schel, and Edward Grefenstette.
\newblock Learning with amigo: Adversarially motivated intrinsic goals.
\newblock \emph{arXiv preprint arXiv:2006.12122}, 2020.

\bibitem[Dennis et~al.(2020)Dennis, Jaques, Vinitsky, Bayen, Russell, Critch,
  and Levine]{dennis2020emergent}
Michael Dennis, Natasha Jaques, Eugene Vinitsky, Alexandre Bayen, Stuart
  Russell, Andrew Critch, and Sergey Levine.
\newblock Emergent complexity and zero-shot transfer via unsupervised
  environment design.
\newblock \emph{Neural Information Processing Systems}, 2020.

\bibitem[Graves et~al.(2017)Graves, Bellemare, Menick, Munos, and
  Kavukcuoglu]{graves2017automated}
Alex Graves, Marc~G Bellemare, Jacob Menick, Remi Munos, and Koray Kavukcuoglu.
\newblock Automated curriculum learning for neural networks.
\newblock \emph{arXiv preprint arXiv:1704.03003}, 2017.

\bibitem[Gur et~al.(2019)Gur, Rueckert, Faust, and
  Hakkani-Tur]{gur2018learning}
Izzeddin Gur, Uli Rueckert, Aleksandra Faust, and Dilek Hakkani-Tur.
\newblock Learning to navigate the web.
\newblock In \emph{ICLR}, 2019.

\bibitem[Jakobi(1997)]{jakobi1997evolutionary}
Nick Jakobi.
\newblock Evolutionary robotics and the radical envelope-of-noise hypothesis.
\newblock \emph{Adaptive behavior}, 6\penalty0 (2):\penalty0 325--368, 1997.

\bibitem[Leibo et~al.(2019)Leibo, Hughes, Lanctot, and
  Graepel]{leibo2019autocurricula}
Joel~Z Leibo, Edward Hughes, Marc Lanctot, and Thore Graepel.
\newblock Autocurricula and the emergence of innovation from social
  interaction: A manifesto for multi-agent intelligence research.
\newblock \emph{arXiv preprint arXiv:1903.00742}, 2019.

\bibitem[Liu et~al.(2018)Liu, Guu, Pasupat, Shi, and
  Liang]{liu2018reinforcement}
Evan~Zheran Liu, Kelvin Guu, Panupong Pasupat, Tianlin Shi, and Percy Liang.
\newblock Reinforcement learning on web interfaces using workflow-guided
  exploration.
\newblock \emph{arXiv preprint arXiv:1802.08802}, 2018.

\bibitem[Matiisen et~al.(2019)Matiisen, Oliver, Cohen, and
  Schulman]{matiisen2019teacher}
Tambet Matiisen, Avital Oliver, Taco Cohen, and John Schulman.
\newblock Teacher-student curriculum learning.
\newblock \emph{IEEE transactions on neural networks and learning systems},
  2019.

\bibitem[Mazumdar et~al.(2019{\natexlab{a}})Mazumdar, Ratliff, Jordan, and
  Sastry]{mazumdar2019policy}
Eric Mazumdar, Lillian~J Ratliff, Michael~I Jordan, and S~Shankar Sastry.
\newblock Policy-gradient algorithms have no guarantees of convergence in
  continuous action and state multi-agent settings.
\newblock \emph{arXiv preprint arXiv:1907.03712}, 2019{\natexlab{a}}.

\bibitem[Mazumdar et~al.(2019{\natexlab{b}})Mazumdar, Jordan, and
  Sastry]{mazumdar2019finding}
Eric~V Mazumdar, Michael~I Jordan, and S~Shankar Sastry.
\newblock On finding local nash equilibria (and only local nash equilibria) in
  zero-sum games.
\newblock \emph{arXiv preprint arXiv:1901.00838}, 2019{\natexlab{b}}.

\bibitem[Portelas et~al.(2020)Portelas, Colas, Hofmann, and
  Oudeyer]{portelas2020teacher}
R{\'e}my Portelas, C{\'e}dric Colas, Katja Hofmann, and Pierre-Yves Oudeyer.
\newblock Teacher algorithms for curriculum learning of deep rl in continuously
  parameterized environments.
\newblock In \emph{Conference on Robot Learning}, pp.\  835--853. PMLR, 2020.

\bibitem[Sadeghi \& Levine(2016)Sadeghi and Levine]{sadeghi2016cad2rl}
Fereshteh Sadeghi and Sergey Levine.
\newblock Cad2rl: Real single-image flight without a single real image.
\newblock \emph{arXiv preprint arXiv:1611.04201}, 2016.

\bibitem[Shi et~al.(2017)Shi, Karpathy, Fan, Hernandez, and
  Liang]{shi2017world}
Tianlin Shi, Andrej Karpathy, Linxi Fan, Jonathan Hernandez, and Percy Liang.
\newblock World of bits: An open-domain platform for web-based agents.
\newblock In \emph{International Conference on Machine Learning}, pp.\
  3135--3144, 2017.

\bibitem[Sukhbaatar et~al.(2017)Sukhbaatar, Lin, Kostrikov, Synnaeve, Szlam,
  and Fergus]{sukhbaatar2017intrinsic}
Sainbayar Sukhbaatar, Zeming Lin, Ilya Kostrikov, Gabriel Synnaeve, Arthur
  Szlam, and Rob Fergus.
\newblock Intrinsic motivation and automatic curricula via asymmetric
  self-play.
\newblock \emph{arXiv preprint arXiv:1703.05407}, 2017.

\bibitem[Tobin et~al.(2017)Tobin, Fong, Ray, Schneider, Zaremba, and
  Abbeel]{tobin2017domain}
Josh Tobin, Rachel Fong, Alex Ray, Jonas Schneider, Wojciech Zaremba, and
  Pieter Abbeel.
\newblock Domain randomization for transferring deep neural networks from
  simulation to the real world.
\newblock In \emph{2017 IEEE/RSJ international conference on intelligent robots
  and systems (IROS)}, pp.\  23--30. IEEE, 2017.

\bibitem[Wang et~al.(2019)Wang, Lehman, Clune, and Stanley]{wang2019paired}
Rui Wang, Joel Lehman, Jeff Clune, and Kenneth~O Stanley.
\newblock Paired open-ended trailblazer (poet): Endlessly generating
  increasingly complex and diverse learning environments and their solutions.
\newblock \emph{arXiv preprint arXiv:1901.01753}, 2019.

\bibitem[Wang et~al.(2020)Wang, Lehman, Rawal, Zhi, Li, Clune, and
  Stanley]{wang2020enhanced}
Rui Wang, Joel Lehman, Aditya Rawal, Jiale Zhi, Yulun Li, Jeff Clune, and
  Kenneth~O Stanley.
\newblock Enhanced poet: Open-ended reinforcement learning through unbounded
  invention of learning challenges and their solutions.
\newblock \emph{arXiv preprint arXiv:2003.08536}, 2020.

\end{thebibliography}
